\documentclass{article}
\usepackage{amsmath,graphicx,mlspconf_modi}

\usepackage{url}
\usepackage{color}
\usepackage{hyperref}
\usepackage{subfig}
\usepackage{listings}
\usepackage{inconsolata}

\definecolor{codegreen}{rgb}{0,0.6,0}
\definecolor{codegray}{rgb}{0.5,0.5,0.5}
\definecolor{codepurple}{rgb}{0.58,0,0.82}
\definecolor{backcolour}{rgb}{0.95,0.95,0.92}
 
\lstdefinestyle{mystyle}{
    backgroundcolor=\color{backcolour},   
    commentstyle=\color{codegreen},
    keywordstyle=\color{magenta},
    numberstyle=\tiny\color{codegray},
    stringstyle=\color{codepurple},
    basicstyle=\footnotesize,
    breakatwhitespace=false,         
    breaklines=true,                 
    captionpos=b,                    
    keepspaces=true,                 
    numbers=left,                    
    numbersep=5pt,                  
    showspaces=false,                
    showstringspaces=false,
    showtabs=false,                  
    tabsize=2
}
 
\begin{document}


\title{Explaining Deep Convolutional Neural Networks on Music Classification}
\name{
  \begin{tabular}{c}
Keunwoo Choi, Gy\"orgy Fazekas, Mark Sandler\sthanks{This paper has been supported by EPSRC Grant EP/L019981/1 and the European Commission H2020 research and innovation grant AudioCommons (688382). Sandler acknowledges the support of the Royal Society as a recipient of a Wolfson Research Merit Award.. This paper is written based on our demonstration, \cite{choiauralisation}.} \\
\end{tabular}
}
\address{
\begin{tabular}{c}
Queen Mary University of London,
London, The United Kingdom \\
\href{mailto:keunwoo.choi@qmul.ac.uk}{\texttt{\{keunwoo.choi, g.fazekas, mark.sandler\}@qmul.ac.uk}}\\
\end{tabular}
}



\maketitle

\begin{abstract}
Deep convolutional neural networks
(CNNs) have been actively adopted in the field of music
information retrieval, e.g. genre classification, mood detection, 
and chord recognition. However, the process 
of learning and prediction is little understood, particularly 
when it is applied to spectrograms. We introduce auralisation 
of a CNN to understand its underlying mechanism, which is 
based on a deconvolution procedure introduced in 
\cite{zeiler2014visualizing}. Auralisation of a CNN is 
converting the learned convolutional features that are obtained from deconvolution
into audio signals.
In the experiments and discussions, we explain
trained features of a 5-layer CNN based on the
deconvolved spectrograms and auralised signals.
The pairwise correlations per layers with varying different
musical attributes are also investigated to understand
the evolution of the learnt features.
It is shown that in the deep layers, the
features are learnt to capture textures, the patterns of continuous distributions, rather than shapes of lines.
\end{abstract}

\section{Introduction}\label{sec:introduction}

In the field of computer vision, deep learning approaches 
became \textit{de facto standard} since convolutional neural 
networks (CNNs) showed break-through results in the ImageNet 
competition in 2012 \cite{krizhevsky2012imagenet}. 
The strength of these approaches comes from the \textit{feature learning} procedure, where
every parameters is learnt to reduce the loss function.

CNN-based approaches have been also adopted in music information retrieval.
For example, 2D convolutions
are performed for music-noise segmentation \cite{parkmusic} and
chord recognition \cite{humphrey2012rethinking} while 1D
(time-axis) convolutions are performed for automatic tagging
in \cite{dieleman2014end}.

The mechanism of learnt filters in CNNs is relatively clear when the
target shapes are known. 
What has not been demystified yet is how CNNs work for tasks such as
mood recognition or genre classification. Those tasks are related
to subjective, perceived impressions, 
whose relation to acoustical properties and whether neural networks models can learn relevant and
optimal representation of sound that help to boost performance in these tasks is an open question.
As a result, researchers currently lack on understanding of
what is learnt by CNNs when CNNs are used in those tasks, even
if it show state-of-the-art performance \cite{dieleman2014end, ullrich2014boundary}.

One effective way to examine CNNs 
was introduced in \cite{zeiler2014visualizing}, where the 
features in deeper levels are visualised by a method called 
\textit{deconvolution}. Deconvolving and un-pooling 
layers enables people to see which part of the input 
image are focused on by each filter. 
However, it does not provide a relevant explanation of CNNs on
music, because the behaviours of CNNs are task-specific and data-dependent.
Unlike visual image recognition tasks, where outlines of images
play an important role, spectrograms mainly consist of continuous,
smooth gradients. There are not only local correlations but also global correlations
such as harmonic structures and rhythmic events. 


In this paper, we introduce the procedure and results of 
deconvolution and auralisation to extend our understanding
of CNNs in music. We not only apply deconvolution to
the spectrogram, but also propose auralisation of the
trained  filters to achieve time-domain reconstruction. In Section \ref{sec:background}, the 
background of CNNs and deconvolution are explained. 
The proposed auralisation method is introduced in Section
 \ref{sec:recon}.  The experiment results are discussed in 
Section \ref{sec:results}. Conclusions are presented in Section \ref{sec:conc}.

\section{Background}\label{sec:background}
\subsection{Visualisation of CNNs}

Multiple convolutional layers lie at the core of a CNN.
The output of a layer (which is called a feature map) is fed into the input of the following layer. 
This stacked structure enables each layer to learn filters in different levels
of the hierarchy. The subsampling layer is also an important part of CNNs.
It resizes the feature maps and let the network see the data in
different scales.
Subsampling is usually
implemented by max-pooling layers, which add location invariances.
The behaviour of a CNN is not deterministic as the operation
of max-pooling varies by input by input. 
This is why analysing the learnt
weights of convolutional layers does not provide satisfying explanations.

\begin{figure}
    \centering
    \includegraphics[width=\columnwidth]{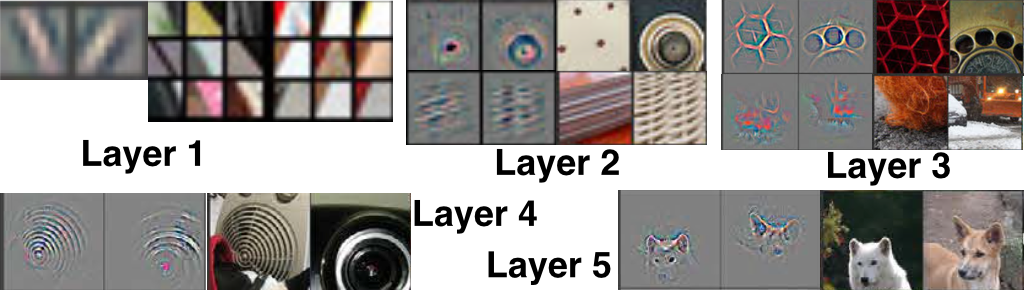}
    \caption{Deconvolution results of CNNs trained for image classification. In each layer, the responses of the filters are shown on the left with gray backgrounds and the corresponding parts from images are shown on the right. Image is courtesy of \cite{zeiler2014visualizing}.}
    \label{fig:conv_example}
\end{figure}

A way to understand a CNN is to visualise the features 
given different inputs. Visualisation of CNNs was introduced 
in \cite{zeiler2014visualizing}, which showed how high-level 
features (postures/objects) are constructed by combining 
low-level features (lines/curves), as illustrated in Figure \ref{fig:conv_example}. 
In the figure, the shapes that features represent evolve.
In the first layers, each feature simply responds to lines with different 
directions. By combining them, the features in the second and 
third layers can capture certain shapes - a circle, textures, 
honeycombs, etc. During this forward path, the features not only 
become more complex but also allow slight variances, and that is
how the similar but different faces of dogs can be recognised
by the same feature in Layer 5 in Figure \ref{fig:conv_example}.
Finally, the features in the
final layer successfully capture the outlines of the target
objects such as cars, dogs, and human faces.

Visualisation of CNNs helps not only to understand the
 process inside the black box model, but also to decide 
hyper-parameters of the networks. For example, redundancy 
or deficiency of the capacity of the networks, which is limited 
by hyper-parameters such as the number of layers and 
filters, can be judged by inspecting the learnt filters. 
Network visualisation provides useful information since fine 
tuning hyper-parameters is a crucial factor in obtaining
cutting-edge performance.

\subsection{Audio and CNNs}
Much research of CNNs on audio signal uses 2D time-frequency 
representations as input data.
Various types of representations have been 
used including Short-time Fourier transform (STFT), 
Mel-spectrogram and constant-Q transform (CQT). CNNs 
show state-of-the-art performance on many tasks 
such as music structure segmentation\cite{ullrich2014boundary}, 
music tagging\cite{dieleman2014end}, and speech/music 
classification\footnote{\url{http://www.music-ir.org/mirex/wiki/2015:Music/Speech_Classification_and_Detection_Results}}. 

Those performances empirically show that CNNs are strong
models to solve many music-related problems. 
Spectrograms match well with the assumption of CNNs from many perspectives.
They are locally correlated, shift/translation invariances
are often required, the output labels may depend on 
local, sparse features \cite{lecun2013deep}.


\section{Auralisation of Learnt Filters}\label{sec:recon}

Although we can obtain spectrograms by deconvolution,
deconvolved spectrograms do not necessarily facilitate an intuitive explanation. 
This is because seeing a spectrogram does not 
necessarily provide clear intuition that is comparable to 
observing an image.

To solve this problem, we propose to reconstruct audio 
signals from deconvolved spectrograms,
which is called \textit{auralisation}. This requires 
an additional stage for inverse-transformation of a 
deconvolved spectrogram. The phase information is provided 
by the phase of the original time-frequency representations, following the generic 
approach in spectrogram-based sound source separation 
algorithms \cite{fitzgerald2012use}. STFT is therefore recommended as it allows us 
to obtain a time-domain signal easily.

Pseudo code of the auralisation is described in Listing 1. Line 1 indicates that
we have a convolutional neural network that is trained for a target task. In line 2-4, an STFT representation of a music signal is provided. Line 5 computes the weights of the neural networks with the input STFT representation and the result is used during the deconvolution of the filters in line 6 (\cite{zeiler2014visualizing} for more details). Line 7-9 shows that the deconvolved filters can be converted into time-domain signals by applying the phase information and inverse STFT.

\begin{lstlisting}[language=Python, caption=A pseudo-code of auralisation procedure]
cnn_model  = train_CNNs(*args) # model
src        = load(wavfile)
SRC        = stft(src)
aSRC, pSRC = SRC.mag, SRC.phase
weights = unpool_info(cnn_model, aSRC)
deconved_imgs = deconv(weights, aSRC)
for img in deconved_imgs:
    signal = inverse_stft(img * pSRC)
    wav_write(signal)
\end{lstlisting} 
\label{list:code}	


\section{Experiments and Discussions} \label{sec:results}

We implemented a CNN-based genre classification algorithm 
using a dataset obtained from \textit{Naver Music} 
\footnote{http://music.naver.com, a Korean music streaming service} 
and based on \textit{Keras} \cite{chollet2015} and \textit{Theano} \cite{2016arXiv160502688short}. 
All audio signal processing was done 
using \textit{librosa} \cite{mcfee2015librosa}. Three genres (\textit{ballad}, 
\textit{dance}, and \textit{hiphop}) were classified using 8,000 songs 
in total.
In order to maximally exploit the data, 10 clips of 4 seconds were extracted for each song, generating 
80,000 data samples by STFT. STFT is computed with 512-point windowed 
Fast Fourier Transform with 50\% hop size and sampling 
rate of 11,025 Hz. 6,600/700/700 songs were designated 
as training/validation/test sets respectively. 

\begin{figure}[t]
    \centering
    \includegraphics[width=0.9\columnwidth]{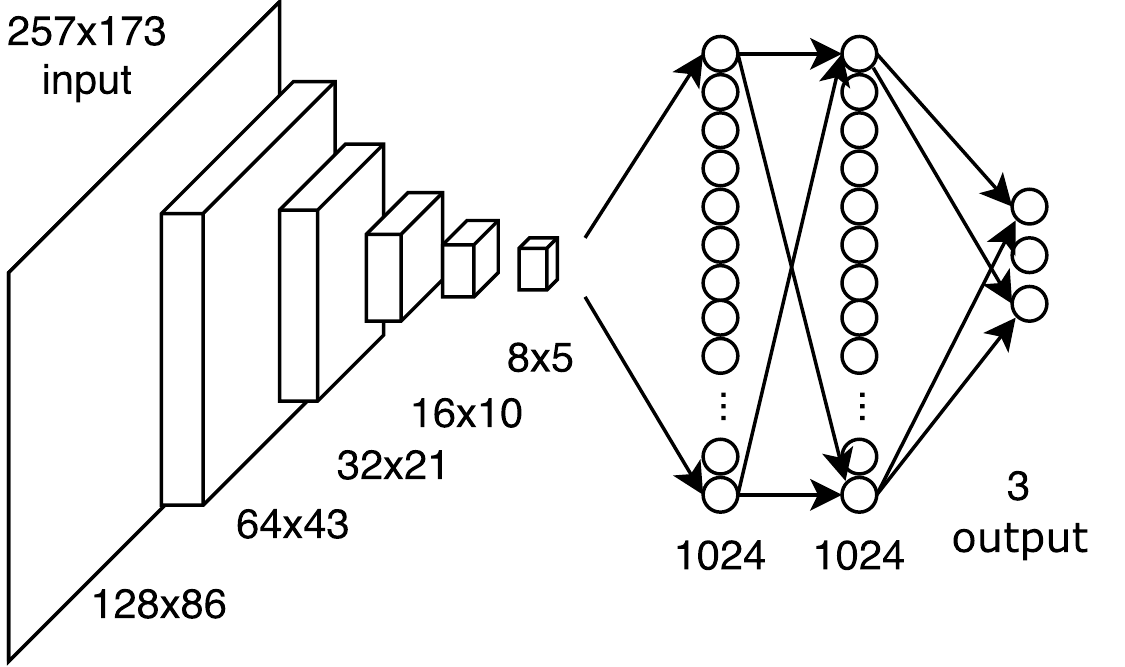} 
    \caption{Block diagram of the trained CNNs. }
    \label{fig:diagram}
\end{figure}

The CNN architecture consists of 5 convolutional layers of 64 feature 
maps and 3-by-3 convolution kernels, max-pooling with 
size and stride of (2,2), and two fully connected layers as 
illustrated in the figure \ref{fig:diagram}. Dropout(0.5) is 
applied to the all convolutional and fully-connected layers to
increases generalisation \cite{srivastava2014dropout}. 
This system showed 75\% of accuracy at the end of training.

\begin{table}[t!]
\begin{center}
\begin{tabular}{c|c|r|r}

Layer & Convolution & Effective width & Effective height\\ \hline
\hline
1 & 3$\times$3 & 93 ms &86 Hz \\ \hline
2 & 3$\times$3 & 162 ms & 151 Hz \\ \hline 
3 & 3$\times$3 & 302 ms & 280 Hz\\ \hline
4 & 3$\times$3 & 580 ms& 538 Hz\\ \hline
5 & 3$\times$3 & 1137 ms& 1270 Hz\\ 
\end{tabular}
\caption{The effective sizes of convolutional kernels}
\label{table:coverages}	
\end{center}
\end{table}


It is noteworthy that although a homogeneous size of convolutional
kernels (3$\times$3) are used, the effective coverages are 
increasing due to the subsampling, as in the table \ref{table:coverages}.

We show two results in the following sections. In Section \ref{exp:music},
the deconvolved spectrograms of selected learnt filters are presented
with discussions. Section \ref{exp:model} shows how the learnt filters
respond by the variations of key, chord, and instrument.

\subsection{Deconvolved results with music excerpts}\label{exp:music}
We deconvolved and auralised the learnt 
features of Four music signals by \textit{Bach}, \textit{Lena Park (Dream)},
 \textit{Toy}, and \textit{Eminem}. Table \ref{table:songs} 
 describes the items. In the following section, several 
 selected examples of deconvolved spectrograms are illustrated
  with descriptions.\footnote{The results are demonstrated 
on-line at \url{http://wp.me/p5CDoD-k9}. An example code of the deconvolution procedure is released at \url{https://github.com/keunwoochoi/Auralisation}} The descriptions are not the
designated goals but interpretations of the features.
 During the overall process, listening to auralised signals helped to identify
pattern in the learnt features.

\begin{table}[t!]
\begin{center}
\begin{tabular}{c|l}

  Name & Summary \\ \hline \hline
  Bach & Classical, piano solo \\ \hline
  Dream & Pop, female vocal, piano, bass guitar \\ \hline
  Toy & Pop, male vocal, drums, piano, bass guitar \\ \hline
  Eminem & Hiphop, male vocal, piano, drums, bass guitar\\ 
\end{tabular}
\caption{Descriptions of the four selected music items}
\label{table:songs}	
\end{center}
\end{table}

\subsubsection{Layer 1}

\begin{figure}[t!]
    \centering
     \includegraphics[width=\columnwidth]{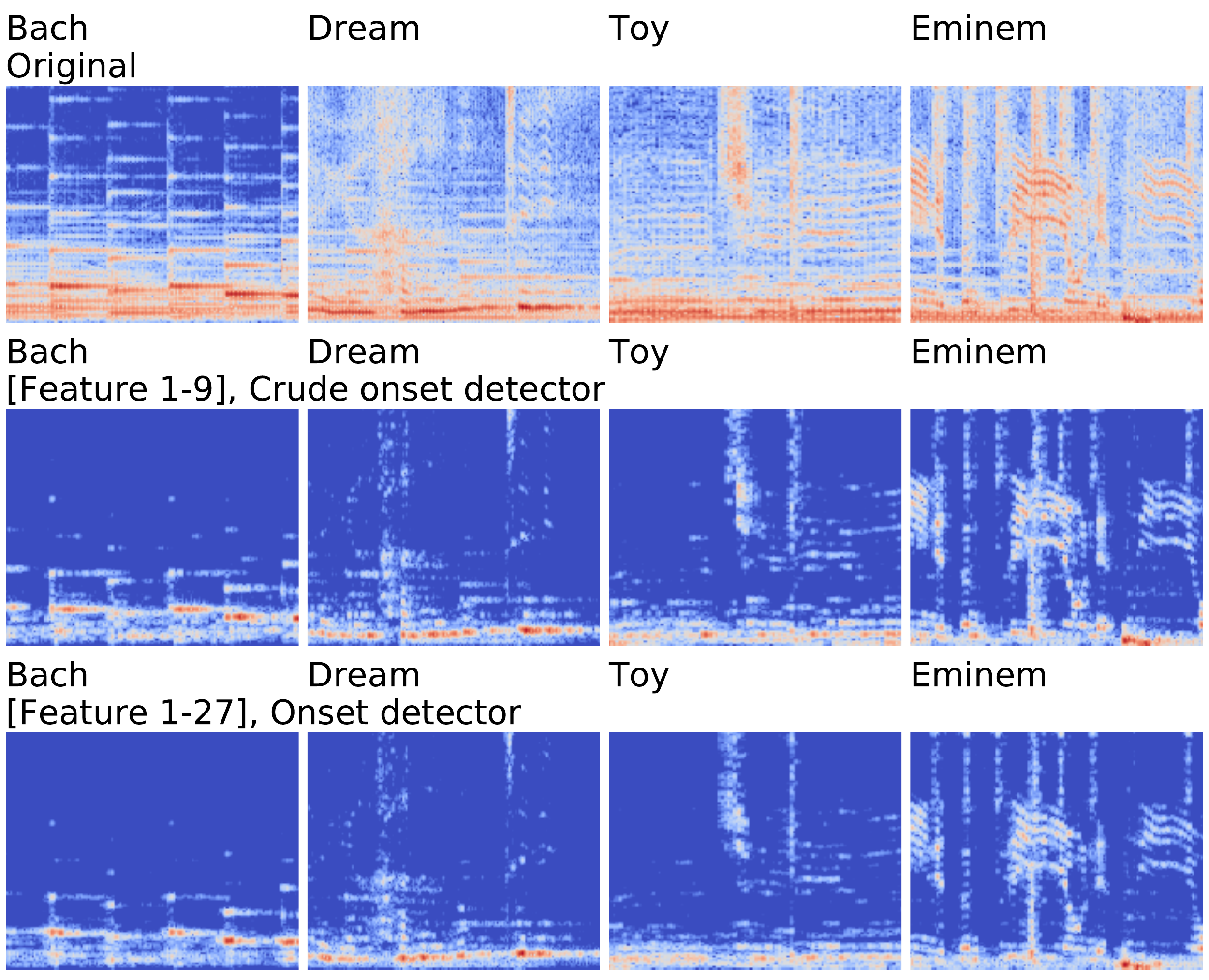}
    \caption{Spectrograms of deconvolved signal in Layer 1}
    \label{fig:layer1}
\end{figure}


\begin{figure}[h!]
    \centering
     \includegraphics[width=\columnwidth]{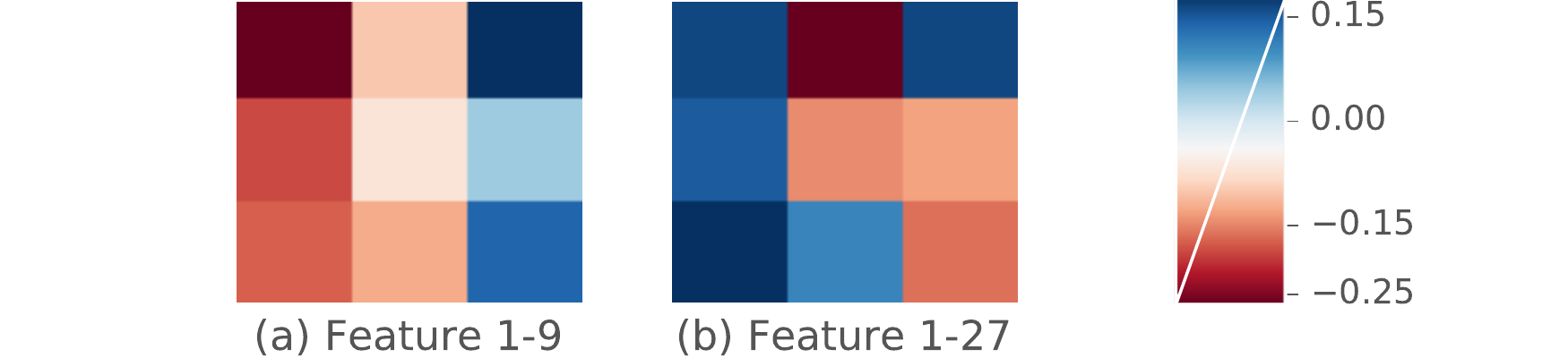}
     \caption{The learnt weights of Features (a) 1-9 and (b) 1-27. The distributions
along rows and columns roughly indicate high-pass filter behaviours along x-axis (time axis) and 
low-pass filter behaviours along y-axis (frequency axis). As a result, they behave as onset detectors.}

    \label{fig:layer1_vals}
\end{figure}


In Layer 1, we selected two features and present their
deconvolved spectrograms as well as the corresponding weights.
Because the weights in the first layer are applied to the input directly without max-pooling,
the mechanism are determined and can be analysed by inspecting the weights regardless of input.
For instance, Feature 1-9 (9th feature in Layer 1) and Feature 1-27, 
works as an onset detector. The learnt weights are shown
in Figure \ref{fig:layer1_vals}. By inspecting the numbers, it can be
easily understood that the network learnt to capture vertical lines.
In spectrograms, vertical lines often corresponds to the
time-frequency signature of percussive instruments.

Many other features showed similar behaviours in Layer 1.
In other words, the features in Layer 1 learn to represent
multiple onset detectors and suppressors with subtle
differences. It is a similar result
to the result that is often obtained in visual image recognition, where CNNs learn line detectors
with various directions (also known as \textit{edge detectors}), which are
combined to create more complex shapes in the second layer. 
With spectrograms, the network focuses on detecting horizontal and vertical
edges rather than diagonal edges. This may be explained by the energy distribution
in spectrograms. Horizontal and vertical lines are main components of 
harmonic and percussive instruments, respectively, while diagonal lines
mainly appear in the case of frequency modulation, which is relatively rare.

\subsubsection{Layer 2}

\begin{figure}[t!]
    \centering
     \includegraphics[width=\columnwidth]{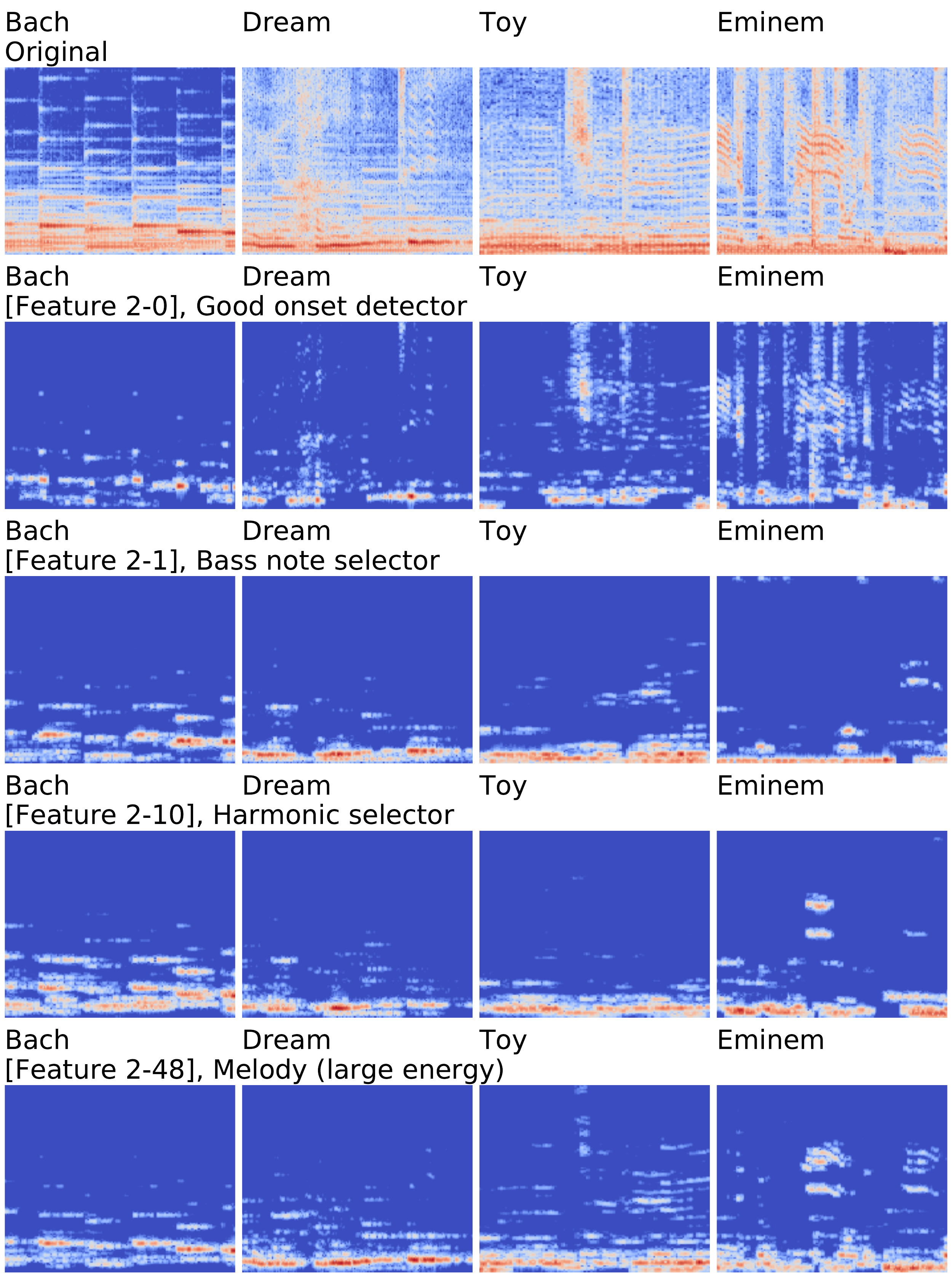}
    \caption{Spectrograms of deconvolved signal in Layer 2}
    \label{fig:layer2}
\end{figure}

Layer 2 shows more evolved, complex features compared to
Layer 1. Feature 2-0 is an advanced (or stricter) onset detectors 
than the onset detectors in Layer 1. 
This improvement can be explained from two perspectives. First, as 
the features in Layer 2 can cover a wider range both in 
time and frequency axis than in layer 1, non-onset parts of signals can be suppressed more effectively. 
Second, the multiple onset detectors in Layer 1
can be combined, enhancing their effects. 

Feature 2-1 (\textit{bass note}), roughly selects the 
lowest fundamental frequencies given harmonic patterns. 
Feature 2-10 behaves as a harmonic component selector, 
excluding the onset parts of the notes. Feature 2-48
is another harmonic component selector with subtle differences.
It behaves as a short melodic fragments 
extractor, presumably by extracting the
most salient harmonic components.

\subsubsection{Layer 3}
 
\begin{figure}[t!]
    \centering
     \includegraphics[width=\columnwidth]{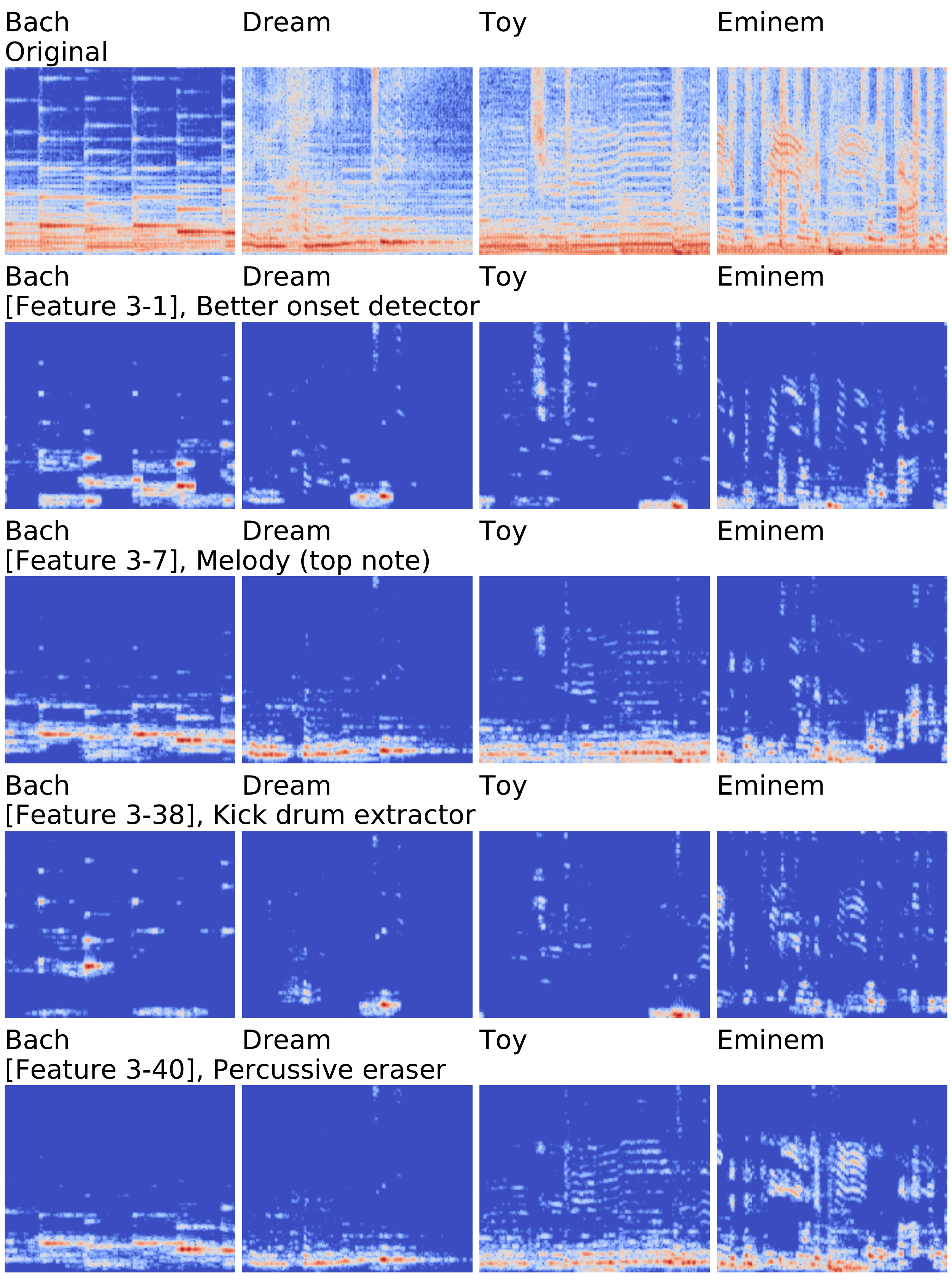}
    \caption{Spectrograms of deconvolved signal in Layer 3}
    \label{fig:layer3}
\end{figure}

The patterns of some features in Layer 3 are similar to that of Layer 2.
However, some of the features in Layer 3 contain higher-level information
e.g. focusing on different instrument classes.

The deconvolved signal from Feature 3-1 consists of onsets of
harmonic instruments, being activated by voices and piano sounds
but not highly activated by hi-hats and snares. The sustain and 
release parts of the envelopes are effectively filtered out in this feature. 
Feature 3-7 is similar to Feature 2-48 but it is
more accurate at selecting the fundamental frequencies of top notes.
Feature 3-38 extracts the 
sounds of kick drum with a very good audio quality. 
Feature 3-40 effectively suppresses transient parts,
resulting softened signals. 

The learnt features imply that the roles of some learnt features
are analogous to tasks such as harmonic-percussive separation,
onset detection, and melody estimation.

\subsubsection{Layer 4}

\begin{figure}[t!]
    \centering
     \includegraphics[width=\columnwidth]{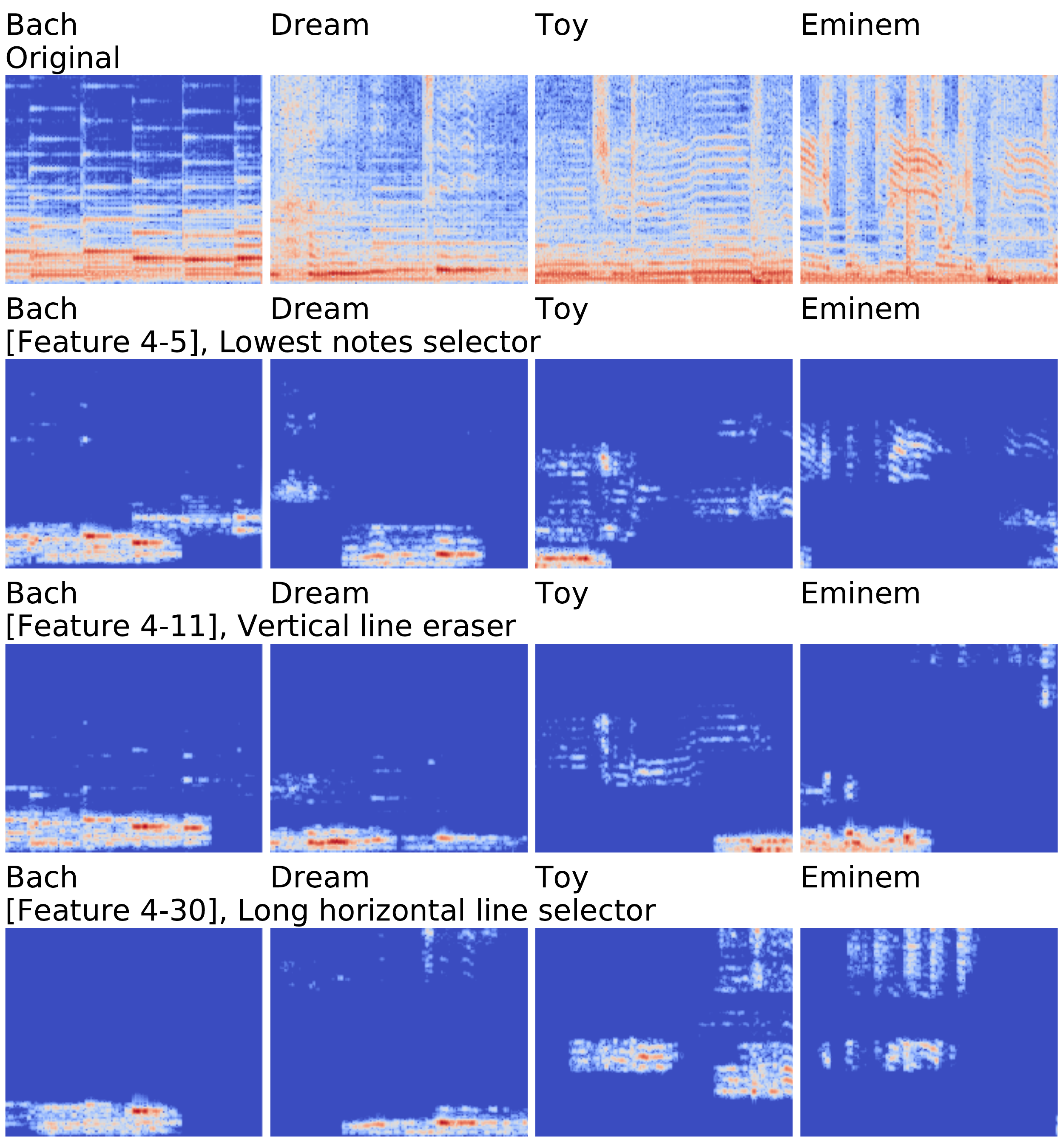}
    \caption{Spectrograms of deconvolved signal in Layer 4}
    \label{fig:layer4}
\end{figure}

Layer 4 is the second last layer of the convolutional layers in the architecture
and expected to represent high-level features. 
In this layer, a convolutional kernel covers a large area (580 ms$\times$538 Hz), which affects
the deconvolved spectrograms. It becomes trickier to name the features by their characteristics, although their activation behaviours are still stable on input data. 
Feature 4-11 removes vertical lines and captures another harmonic texture. 
Although the coverages of the kernels increase, the features in Layer 4 try to find
patterns more precisely rather than simply respond to common shapes such as edges.  
As a result, the activations become more sparse (local) because a feature responds 
only if a certain pattern -- that matches the learnt features -- exists, ignoring irrelevant patterns.

\subsubsection{Layer 5}

This is the final layer of the convolutional layers, and therefore it  
represents the highest-level features among all 
the learnt features. High-level features respond to latent and abstract concepts, which
makes it more difficult to understand 
by either listening to auralised signals or seeing deconvolved spectrograms. 
Feature 5-11, 5-15, and 5-33 are therefore named as textures. 
Feature 5-56, harmo-rhythmic texture, 
is activated almost only if strong percussive 
instruments \textit{and} harmonic patterns overlap. Feature 5-33
is completely inactivated with the fourth excerpt, which 
is the only Hip-Hop music, suggesting it may be useful for 
classification of Hip-Hop.

\begin{figure}[t!]
    \centering
     \includegraphics[width=\columnwidth]{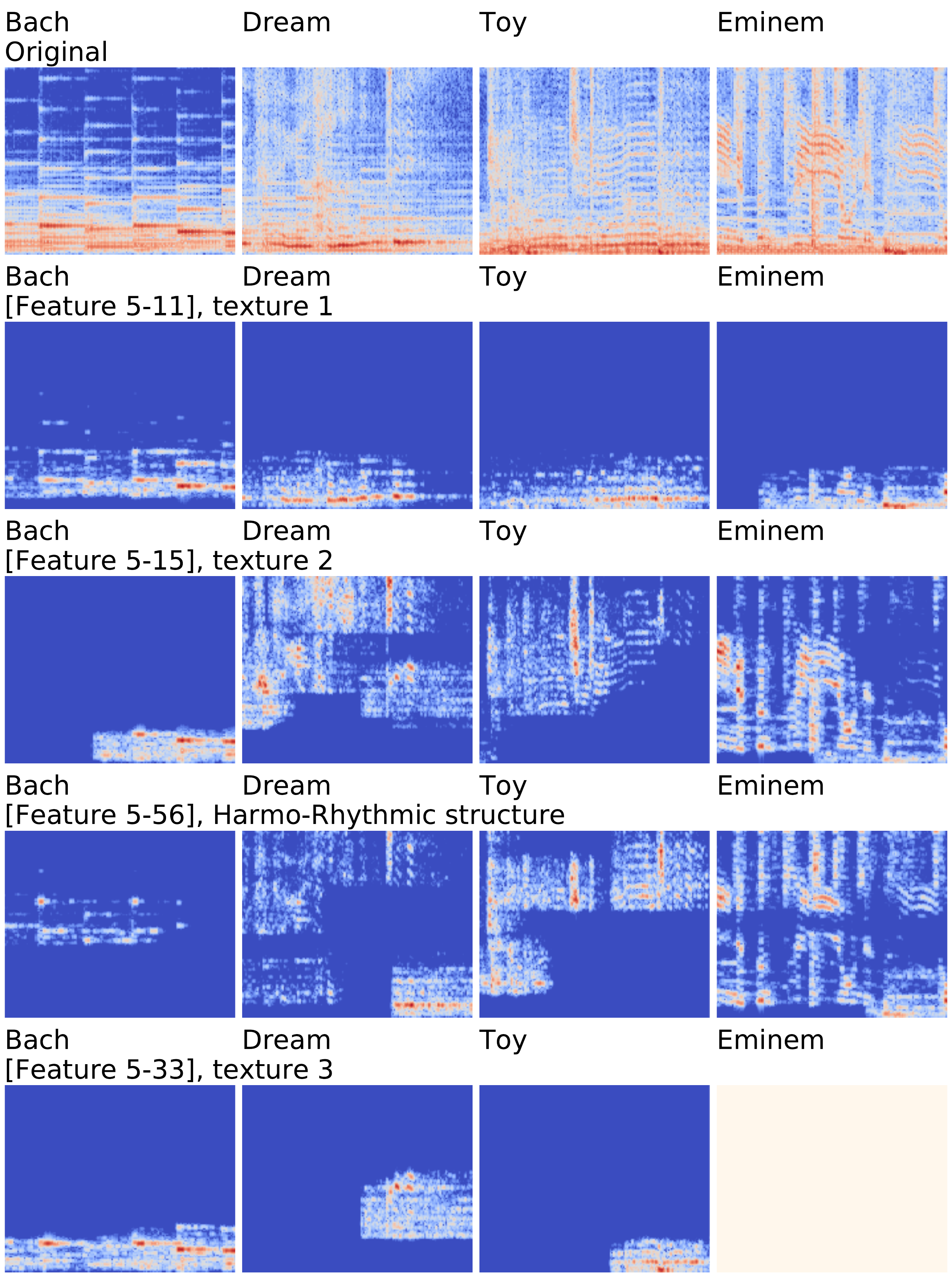}
    \caption{Spectrograms of deconvolved signal in Layer 5}
    \label{fig:layer5}
\end{figure}

\subsection{Feature responses by attributes}\label{exp:model}
\begin{figure}[t!]
    \centering
     \includegraphics[width=\columnwidth]{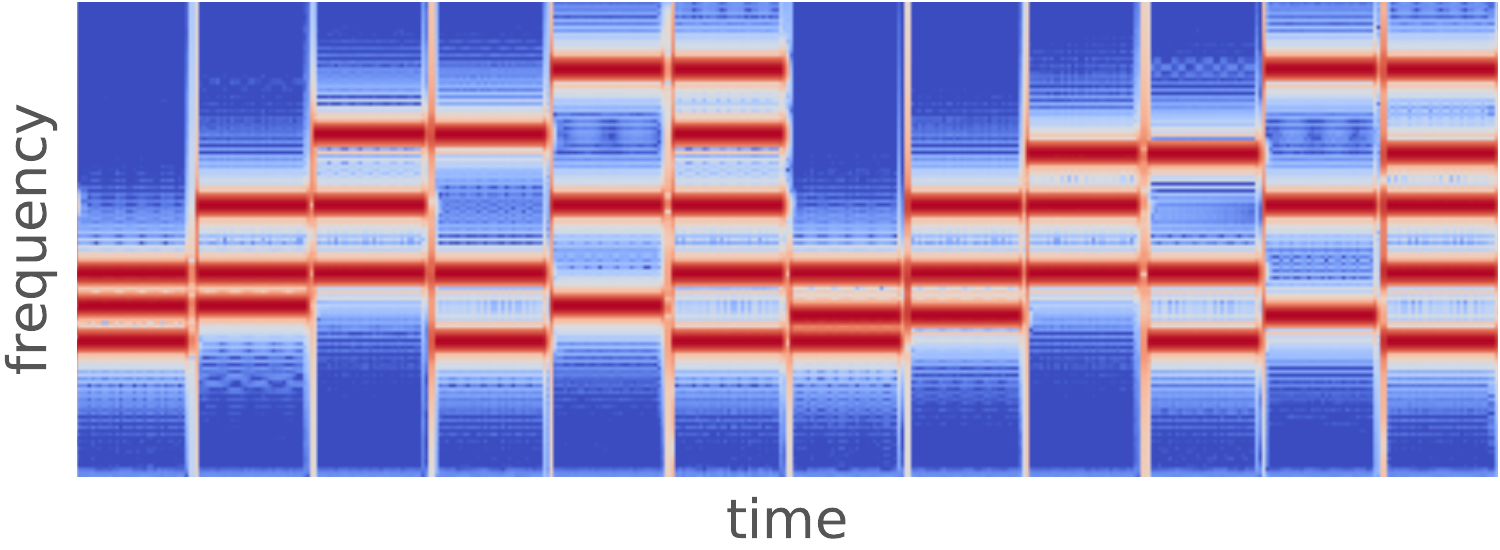}
    \caption{A spectrogram of an example of the model signals -- at the key of G4, an instrument of pure sine, and the chord of six major positions (first half) and six minor positions (second half). High-frequency ranges are trimmed for better frequency resolution.}
    \label{fig:model_signal}
\end{figure}

In this experiment, a set of \textit{model} signals are created. 
Model signals are simplified music excerpts, each of which
consists of a harmonic \textit{instrument} playing various \textit{chords} at different \textit{keys}.
In total, 7 instruments (\textit{pure sine, strings, acoustic guitar, saxophone, piano, electric guitar}) $\times$
8 chord types (\textit{intervals, major, minor, sus4, dominant7, major7, minor7, diminished}) $\times$
4 keys (\textit{Eb2, Bb2, A3, G4}) are combined, resulting in 224 model signals.
Figure \ref{fig:model_signal} shows the spectrogram of one of the model signals.

\begin{figure}[b!]
    \centering
     \includegraphics[width=\columnwidth]{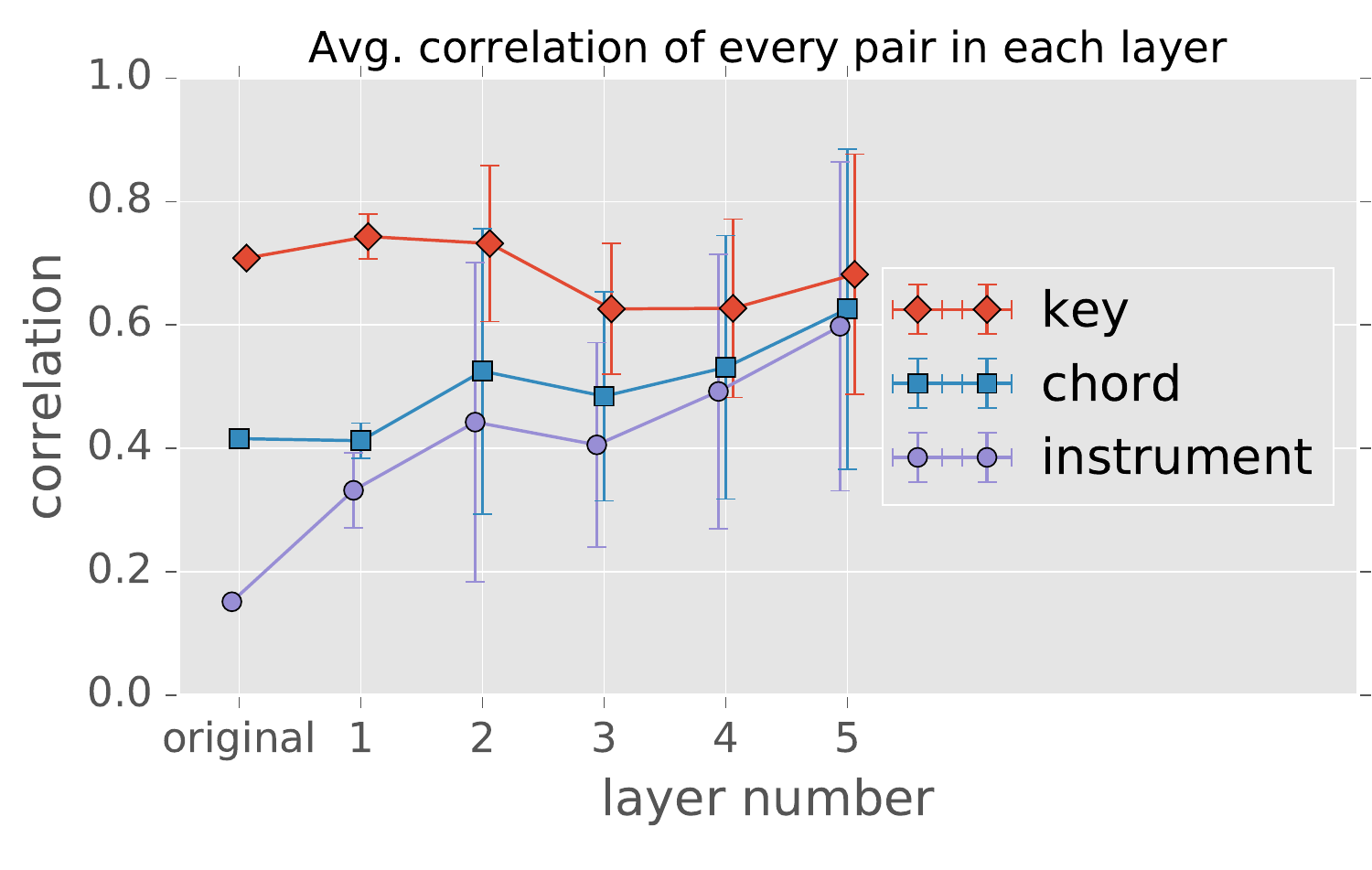}
    \caption{Average correlation of every pairs in each layer by the variations of key, chord, and instrument. Error bars refer to
    the corresponding standard deviations.}
    \label{fig:avg_corr}
\end{figure}

All the model signals are fed into the trained CNN. Then, all the
learnt features are deconvolved, resulting in 64 spectrograms 
per layer and per model signal. In other words, there are 224 spectrograms
for each feature. If we compute the average correlation of the 6 pairs of keys
(\{Eb2, Bb2\}, \{Eb2, A3\}, \{Eb2, G4\}, \{Bb2, A3\}, \{Bb2, G4\}, \{A3, G4\}) from the features in Layer 1 with fixing
the chord and instrument, 
we can see how sensitive (or not robust) the CNN is to key changes in Layer 1. 
The robustness of the other layers and to chord or instrument can be computed in the same manner.
We computed the average of this correlation for all features, every
pairs of key, chord, and instrument, and per layer. The result is plotted in
Figure \ref{fig:avg_corr}. 

According to Figure \ref{fig:avg_corr}, key variations have the smallest effect on the CNN for genre classification 
over the whole network. It agrees well with general understanding of genre, in which
key does not play an important role.
The low correlation with chord type variations in Layer 1 indicates the features
in early layers are affected.
However, the effect decreases as progressing towards deeper layers. At Layer 5, the
chord pairs become more correlated then they do in the early layers,
which means more robusness.
The CNN is the most sensitive to the instrument variations at Layer 1.
Considering the simple features in Layer 1, the different onset and harmonic
patterns by instruments may contribute this.
However, it becomes more robust in the deeper layer. 
At Layer 5, instruments are slightly less correlated than chords are.

To summarise, all three variations show similar average correlations
in Layer 5
, indicating the high-level features
that CNN learnt to classifier genre are robust to the variations of key, chord, and
instrument.

\section{Conclusions}\label{sec:conc}
We introduced auralisation of CNNs, which is an extension of CNNs visualisation. 
This is done by inverse-transformation of deconvolved spectrograms to obtain time-domain audio signals. 
Listening to the audio signal enables researchers to understand the mechanism of CNNs that are trained with audio signals. 
In the experiments, we trained a 5-layer CNN to classify genres. Selected learnt features are reported with interpretations from musical and music information aspects. The comparison of correlations of feature responses showed how the features evolve and become more invariant to the chord and instrument variations.
Further research will include computational analysis of learnt features.

\bibliographystyle{IEEEbib}
\bibliography{auralisation_arxiv}

\begin{thebibliography}{10}

\bibitem{choiauralisation}
Keunwoo Choi, George Fazekas, Mark Sandler, and Jeonghee Kim,
\newblock ``Auralisation of deep convolutional neural networks: Listening to
  learned features,''
\newblock {\em ISMIR late-breaking session}, 2015.

\bibitem{zeiler2014visualizing}
Matthew~D Zeiler and Rob Fergus,
\newblock ``Visualizing and understanding convolutional networks,''
\newblock in {\em Computer Vision--ECCV 2014}, pp. 818--833. Springer, 2014.

\bibitem{krizhevsky2012imagenet}
Alex Krizhevsky, Ilya Sutskever, and Geoffrey~E Hinton,
\newblock ``Imagenet classification with deep convolutional neural networks,''
\newblock in {\em Advances in neural information processing systems}, 2012, pp.
  1097--1105.

\bibitem{parkmusic}
Taejin Park and Taejin Lee,
\newblock ``Music-noise segmentation in spectrotemporal domain using
  convolutional neural networks,''
\newblock {\em ISMIR late-breaking session}, 2015.

\bibitem{humphrey2012rethinking}
Eric~J Humphrey and Juan~P Bello,
\newblock ``Rethinking automatic chord recognition with convolutional neural
  networks,''
\newblock in {\em Machine Learning and Applications (ICMLA), International
  Conference on}. IEEE, 2012.

\bibitem{dieleman2014end}
Sander Dieleman and Benjamin Schrauwen,
\newblock ``End-to-end learning for music audio,''
\newblock in {\em Acoustics, Speech and Signal Processing (ICASSP), 2014 IEEE
  International Conference on}. IEEE, 2014, pp. 6964--6968.

\bibitem{ullrich2014boundary}
Karen Ullrich, Jan Schl{\"u}ter, and Thomas Grill,
\newblock ``Boundary detection in music structure analysis using convolutional
  neural networks,''
\newblock in {\em Proceedings of the 15th International Society for Music
  Information Retrieval Conference (ISMIR 2014), Taipei, Taiwan}, 2014.

\bibitem{lecun2013deep}
Yann LeCun and M~Ranzato,
\newblock ``Deep learning tutorial,''
\newblock in {\em Tutorials in International Conference on Machine Learning
  (ICML13), Citeseer}. Citeseer, 2013.

\bibitem{fitzgerald2012use}
Derry Fitzgerald and Rajesh Jaiswal,
\newblock ``On the use of masking filters in sound source separation,''
\newblock {\em Int. Conference on Digital Audio Effects (DAFx-12)}, 2012.

\bibitem{chollet2015}
Fran{\c c}ois Chollet,
\newblock ``Keras: Deep learning library for theano and tensorflow,''
  https://github.com/fchollet/keras, 2015.

\bibitem{2016arXiv160502688short}
{Theano Development Team},
\newblock ``{Theano: A {Python} framework for fast computation of mathematical
  expressions},''
\newblock {\em arXiv e-prints}, vol. abs/1605.02688, May 2016.

\bibitem{mcfee2015librosa}
Brian McFee, Colin Raffel, Dawen Liang, Daniel~PW Ellis, Matt McVicar, Eric
  Battenberg, and Oriol Nieto,
\newblock ``librosa: Audio and music signal analysis in python,''
\newblock in {\em Proceedings of the 14th Python in Science Conference}, 2015.

\bibitem{srivastava2014dropout}
Nitish Srivastava, Geoffrey Hinton, Alex Krizhevsky, Ilya Sutskever, and Ruslan
  Salakhutdinov,
\newblock ``Dropout: A simple way to prevent neural networks from
  overfitting,''
\newblock {\em The Journal of Machine Learning Research}, vol. 15, no. 1, pp.
  1929--1958, 2014.

\end{thebibliography}
\end{document}